\documentclass[10pt,twocolumn,letterpaper]{article}

\usepackage{cvpr}
\usepackage{times}
\usepackage{epsfig}
\usepackage{graphicx}
\usepackage{amsmath}
\usepackage{amssymb}
\usepackage{color}

\usepackage{amsfonts}       
\usepackage{multirow}
\usepackage{mydefs}
\usepackage{pifont}
\usepackage{graphbox} 
\usepackage{flushend}

\usepackage[pagebackref=true,breaklinks=true,letterpaper=true,colorlinks,citecolor=blue,linkcolor=blue,bookmarks=false]{hyperref}
\usepackage{booktabs}
\usepackage[]{subcaption}
\usepackage{caption}
\usepackage{microtype}
\usepackage{authblk}
\usepackage[titletoc]{appendix}

\cvprfinalcopy 

\def\cvprPaperID{1768} 

\newcommand{\todo}[1]{}
\renewcommand{\todo}[1]{{\color{red} TODO: {#1}}}

\ifcvprfinal\fi
\begin{document}

\title{SoDA: Multi-Object Tracking with Soft Data Association}

\author[1]{Wei-Chih Hung}
\author[1]{Henrik Kretzschmar}
\author[2]{Tsung-Yi Lin}
\author[1]{Yuning Chai}
\author[1]{Ruichi Yu}
\author[2,3]{Ming-Hsuan Yang}
\author[1]{Dragomir Anguelov}
\makeatletter 
\renewcommand\AB@affilsepx{\quad \protect\Affilfont} 
\makeatother
\affil[1]{Waymo LLC}
\affil[2]{Google LLC}
\affil[3]{UC Merced}

\setlength{\fboxsep}{0pt}
\maketitle



\begin{abstract}

Robust multi-object tracking (MOT) is a prerequisite for a safe deployment of self-driving cars.
Tracking objects, however, remains a highly challenging problem, especially in cluttered autonomous driving scenes in which objects tend to interact with each other in complex ways and frequently get occluded.
We propose a novel approach to MOT that uses attention to compute track embeddings that encode the spatiotemporal dependencies between observed objects.
This attention measurement encoding allows our model to relax hard data associations, which may lead to unrecoverable errors.
Instead, our model aggregates information from all object detections via soft data associations.
The resulting latent space representation allows our model to learn to reason about occlusions in a holistic data-driven way and maintain track estimates for objects even when they are occluded.
Our experimental results on the Waymo Open Dataset suggest that our approach leverages modern large-scale datasets and performs favorably compared to the state of the art in visual multi-object tracking.

\end{abstract}


\section{Introduction}

\begin{figure}[t]
  \centering
  \includegraphics[width=1.0\linewidth]{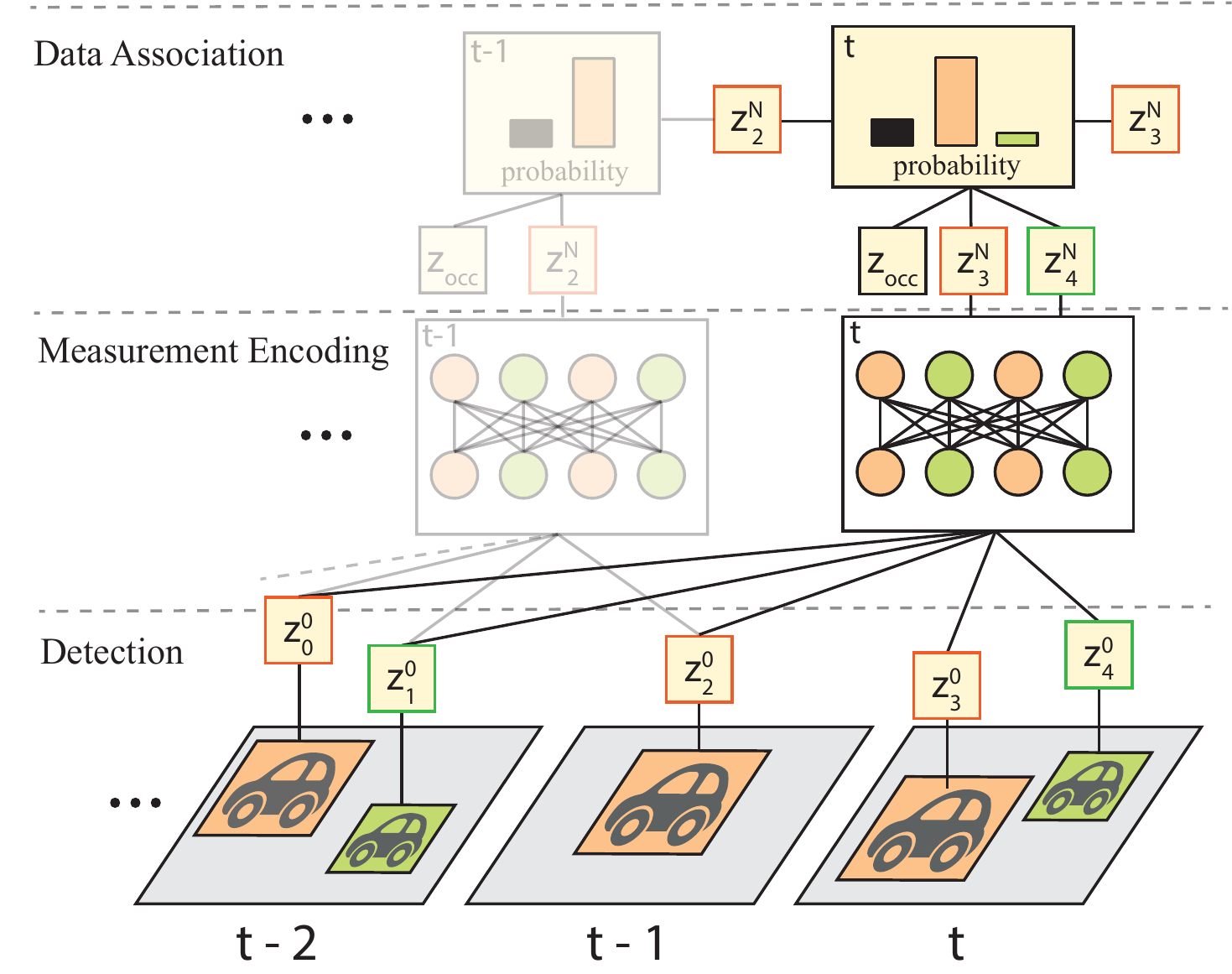}
  \caption{\small\textbf{Overview.}
     We propose a novel approach to multi-object tracking.
  Given object detections based on the measurements $z^0$, our model encodes spatiotemporal context information for each measurement with $N$~self-attention layers, resulting in features~$z^N$ that learn from soft association values, which do not rely on hard-associated tracks. 
  Based on the aggregated features, the model then predicts a probability distribution for each track that captures soft data associations and a latent state $z_{\text{occ}}$, which indicates that the track is occluded.
  }-
  \label{fig:overview}
\end{figure}



Being able to detect and track multiple moving objects in crowded environments simultaneously is a prerequisite for a safe deployment of self-driving cars~\cite{frossard2018end,hu2019joint,luo2018fast}. Despite remarkable advances in object detection in recent years~\cite{ren2015faster,liu2016ssd,lin2017focal}, multi-object tracking~(MOT) in complex scenes remains a highly challenging problem. As the objects move around the scene, they may interact with each other in complex ways that are hard to predict. Furthermore, the objects may frequently occlude each other, which may result in missed detections.
Finally, changes in object appearance, as well as similar object appearances, may make it difficult to recognize previously seen objects.
In the commonly used ``tracking-by-detection'' paradigm, a tracker fuses detections to produce object tracks that are consistent over time. A key challenge, therefore, is to associate incoming detections of previously observed objects with the corresponding existing tracks. 

\begin{figure*}[t]
  \centering
  \includegraphics[width=0.9\linewidth]{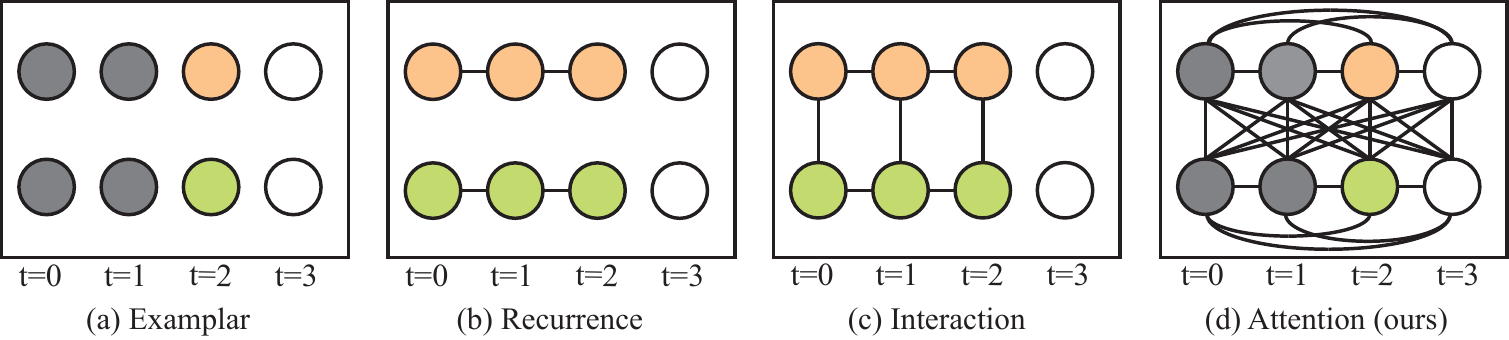}
  \caption{\small\textbf{Comparison of encoding methods.}
  We illustrate how tracking methods use context and history. 
  The red and orange circles represent detections associated with tracks. The white circles represent incoming detections that are yet to be associated.
  (a) Consider how similar an incoming detection is to the latest detection associated with each track.
  (b) Aggregate information from all detections that are associated with each track via hard data association.
  (c) Share information between tracks.
  (d) Aggregate information from all detections to leverage the spatiotemporal context without committing to any hard data associations.
  }
  \label{fig:encoding}
\end{figure*}

Data association in most existing methods is based on similarity scores computed between the detections and the existing tracks. The representation of existing tracks can rely on the last detections~\cite{Wojke17} or it can be aggregated from historical detections associated with the tracks~\cite{Milan17,Sadeghian17}. Figure~\ref{fig:encoding} groups approaches to multi-object tracking by how they leverage context information.

One common assumption made by the above-mentioned methods is that their track state estimates updated by only the hard-associated detections capture all information relevant for subsequent tracking decisions. This assumption, however, is suboptimal as it collapses multiple data association hypotheses into a single mode, ignoring the contextual information from other unselected association candidates and the effects of possible association errors. The error introduced by a single incorrect association may propagate and substantially affect subsequent data associations, especially when the input detections are noisy. 

To relax the hard data association constraint, some methods use multi-hypothesis tracking~\cite{cox1996efficient,Kim18,reid1979algorithm}, and others simply pick the top associated candidates by applying a threshold to the similarities. This, however, requires complex heuristics and hyper-parameter tuning and may easily overfit to certain scenes.
Instead, a unified and data-driven approach may be able to implicitly aggregate information from all candidate detections across frames to update track states without committing to any hard association. Such a soft data association framework may be able to learn long-term and highly interactive relationships between detections and tracks from large-scale datasets without cumbersome heuristics and hyper-parameters.

%

%
%

Inspired by recent advances in natural language processing~\cite{Dai19,Dehghani18,Lee19,Shaw18,Vaswani17}, we formulate MOT as a sequence-to-sequence problem using attention models, where the input of the model is a series of frames with detections, and the output consists of the trajectories of the detected objects.
Specifically, we propose an attention measurement encoding to compute track embeddings that reason about the spatiotemporal dependencies between all objects, as observed by all detections in a given temporal window.
This allows us to avoid hard data association, which is prone to error propagation and may lead to unrecoverable states.
Instead, our method maintains a latent space representation that aggregates information from soft data associations. 
The soft associations further allow us to explicitly reason about occlusions in a data-driven way. Most existing methods determine the associations based on the similarity between an individual pair of a detection and a track. However, picking a single threshold to determine if an object is occluded without context information is challenging.
The track embeddings computed by our method, on the other hand, contain rich contextual information for effective occlusion reasoning.

To evaluate the effectiveness of the proposed approach, we present an extensive ablation study on the Waymo Open Dataset~\cite{waymo_open_dataset} as well as other benchmark datasets. 
The results suggest that our method performs favorably when compared to the state of the art in visual multi-object tracking.
\section{Related Work}
\label{sec:related_work}

\PAR{Multi-Object Tracking.}
Most existing approaches to multi-object tracking, including this one, follow the ``tracking-by-detection'' paradigm.
%
%
Offline methods~\cite{Schulter17,Son17,tang2015subgraph,tang2016multi,Tang17,wang2015tracking,wu2012coupling,zamir2012gmcp,Zhang08} process an entire video in a batch fashion. However, these methods are not applicable to most online application. An autonomous vehicle, for instance, must predict the states of objects immediately when new detections become available.
Most recent approaches to multi-object tracking are therefore online methods that do not depend on future frames~\cite{Bergmann19,Bewley16,Chu17,Kim18,Maksai19,Sadeghian17,Voigtlaender19,Wojke17,Xu19,Zhu18}.
Most online methods estimate similarity scores between the detections and the existing tracks based on various cues, such as predicted bounding boxes~\cite{Bewley16} and appearance similarity~\cite{leal2016learning}.
While some methods~\cite{Bewley16,leal2016learning} only take the last detection corresponding to a track into account, some techniques aggregate temporal information into a track history. For instance, DEEP~Sort~\cite{Wojke17} computes the maximum similarity between the detection and any detections that were previously associated with the track. Other methods rely on recurrent neural networks to accumulate temporal information~\cite{Chu17,Kim18,Milan17,Sadeghian17,Zhu18}. Sadeghian~\etal~\cite{Sadeghian17} apply several recurrent neural networks to learn per-track motion cues, interaction cues, and appearance cues. As opposed to inferring interactions directly from the tracks, their method, however, relies on a rasterized occupancy map as a proxy for inferring interaction cues. 
%
For each incoming detection, the recurrence methods update an internal representation that corresponds to the matching tracks. Here, however, an erroneous data association can cause the internal representation to end up in a bad state, which the method may not be able to recover from.
%
To tackle this issue, we propose to break up the causality between spatiotemporal information aggregation and data association. We encode the detections along with soft attentions to all the other detections in a fixed temporal window into a latent space representation without committing to any hard data associations.

\PAR{Occlusion Reasoning.}
A key challenge in multi-object tracking is to robustly track objects even when they cannot be observed due to occlusions or false negatives of the trained object detector. Even though some methods aim to recover detections by applying a single object tracker~\cite{Zhu18} or a regression head in detector~\cite{Bergmann19}, the object may still be missed during full occlusions. Most online trackers, therefore, predict an occlusion whenever the similarity scores are below a fine-tuned threshold and adopt the buffer-and-recover mechanism~\cite{Kim18} by extrapolating the state with motion model, or by relying on appearance re-identification~\cite{Wojke17}.
Predefined motion models, however, often cannot accurately predict target positions during occlusion, especially in complex interactive scenes.
Choosing a single hand-crafted similarity threshold to predict occlusions in a variety of complex scenes may be suboptimal. We therefore directly learn to explicitly predict occlusions without any hand-crafted geometry or appearance similarity constraints by introducing an occlusion state in a latent representation.
The virtual candidates proposed by FAMNet~\cite{chu2019famnet} are similar to our explicit occlusion reasoning. However, FAMNet chooses the locations of the virtual candidates by using heuristics, while our approach learns the occlusion embedding without any hand-crafted rules.

\PAR{Attention Networks.}
Transformer networks~\cite{Dai19,Dehghani18,Lee19,Shaw18,Vaswani17} and other graph neural networks~\cite{Battaglia18,Vinyals15,Wu19} have recently gained momentum. In addition to successful applications to natural language processing formulated as sequence-to-sequence problems, there is evidence that suggests that Transformer networks can capture long-range dependencies and interactions between agents~\cite{Sun19,Wang_2018_CVPR}.
We formulate multi-object tracking as a sequence-to-sequence problem, where the input consists of sequences of frames with detections, and the output consists of sequences that may have different start and end times.
Other recent methods have adopted attention mechanisms to contextual information~\cite{Chu17,Xu19,Zhu18}. However, these techniques encode information conditioned on trajectories formed by hard data associations.
We propose attention measurement encoding to aggregate spatiotemporal information without conditioning on past trajectory estimates, allowing our method to gracefully handle erroneous data associations.

\section{Track with Soft Data Association}

We adopt the tracking-by-detection paradigm, where a tracker fuses object detections to produce object tracks that are consistent over time.
We propose an attention measurement encoding mechanism to aggregate the spatiotemporal context into a latent space representation without committing to hard data associations. We further propose an attention association mechanism that explicitly reasons about occlusions without any predefined similarity thresholds. 
%
%
%
%
%
The proposed attention measurement encoding processes the detections in a temporal window consisting of object measurements, such as position and appearance, and predicts a fixed-dimensional feature vector for each detection.
The output features are then passed to the attention data association to form target trajectories.
%
See Figure~\ref{fig:overview} for an overview of our approach.
We will explain the two proposed modules in detail in the rest of this section.

%
%

%
%

\subsection{Attention Measurement Encoding}

Most existing tracking methods associate incoming detections in a pairwise fashion with object states predicted by a simple motion model, such as a constant velocity model, using a Kalman filter or recurrent neural networks.
Recent work, however, has demonstrated that aggregating temporal information, as well as context information, may improve multi-object tracking by exploiting higher-order information in addition to pairwise similarities between detections~\cite{Sadeghian17}.
%
%
%
%
We propose to leverage the spatiotemporal dependencies by using an attention measurement encoding mechanism that applies self-attention layers to soft associated detection measurements. This enables us to implicitly learn motion and context models without any hard data associations.

\PAR{Attention with Transformer Network.}
For each new detection obtained at time~$t$ comprising raw measurements~$x_{t,i}$, a feed-forward neural network computes a measurement embedding~$z^0_{t,i}$. This network consists of two fully-connected layers, followed by a normalization layer~\cite{ba2016layer}.
The two fully-connected layers use ReLU and linear activation functions, respectively.
%
%
We then apply stacked attention measurement encoding layers on top of the measurement features~$z^0$ to encode the spatiotemporal context information from each detection without any hard data associations.
Considering detections at time~$t$, our goal is to encode the information from the detections obtained in the past $L_{enc}$~frames, i.e., ${\{z_{\tau,i} | \forall i, \tau \in [t-L_{enc}, t]\}}$.
Following the Transformer architecture~\cite{Vaswani17}, an attention measurement encoding layer consists of two sub-layers, where the first is a self-attention layer, and the second is a point-wise feed forward network.
Both sub-layers follow the ``Add-\&-Norm`` structure adopted by \cite{Vaswani17}, where the output of each layer is used as the input to the following layer by layer normalization~\cite{ba2016layer}, i.e., $z^o = \text{LayerNorm}(F(z^i) + z^i)$.
In the self-attention sub-layer, a measurement embedding feature is updated with scaled dot-product attention, leading to
\begin{equation}
\label{eqn:attention}
    z^o_i = \sum_j \text{softmax}(\dfrac{Q_iK_j^{\top}}{\sqrt{d_k}}) V_i,
\end{equation}
where $Q_i = \mathbf{W}_q z_i$, $K_i = \mathbf{W}_k z_i$ and $V_i = \mathbf{W}_v z_i$ are the query, key, value features obtained by applying a linear transformation on an embedding feature~$z_i$ with size $d_k$, and $j$ denotes the index of all detections in previous $L_{enc}$ frames.
This means that each measurement embedding will be updated by using a weighted sum of past measurements.
After self-attention, we apply layer normalization followed by a feed-forward network with a fully-connected layer and ``Add-\&-Norm''.
Overall, the attention measurement encoding mechanism consists of $N_{enc}$ such attention layers to increase the capacity of the model.

\PAR{Relative Time Encoding.}
Self-attention networks are unordered. It is therefore important to properly encode the temporal information. One way to do this is to have the attention value $A_{i,j} = Q_iK_j^{\top}$ in Equation~\ref{eqn:attention} consider the temporal difference, $(t_i-t_j)$, between $z_i$ and~$z_j$.
In this work, we employ the relative encoding proposed by Transformer-XL~\cite{Dai19}, leading to
\begin{equation}
    A_{i,j} = Q_iK_j^{\top} + Q_iR_{i-j}^{\top} + u K_j^{\top} + v R_{i-j}^{\top}, 
\end{equation}
where $R_k \in \mathcal{R}^{d_k}, k \in [-L_{enc}, L_{enc}]$ is a learned relative attention feature, and $u,v \in \mathcal{R}^{d_k}$ are bias terms for the measurement attention and the relative attention.
Note that, in contrast to our work, the Transformer-XL obtains $R_k$ by applying the linear transform to the sinusoidal positional encoding proposed by the original formulation of the Transformer~\cite{Vaswani17}.
The reason for this modification is that we found that using the sinusoidal positional encoding often leads to non-convergence, whereas directly learning~$R_k$ yields superior performance.
Figure~\ref{fig:encoding} illustrates how our method differs from existing methods in terms of how contextual information flows while tracking objects.


\begin{figure}[t]
  \centering
  \includegraphics[width=1.0\linewidth]{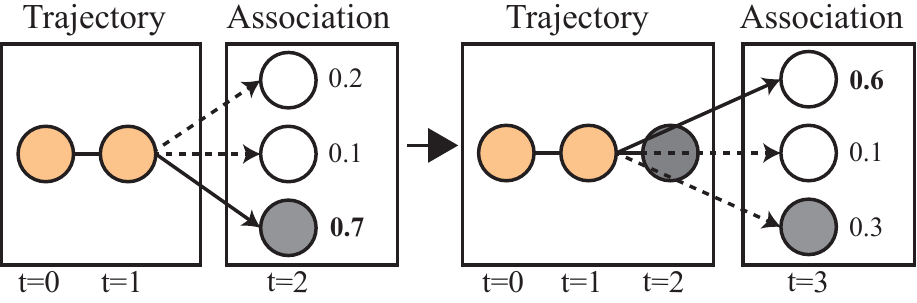}
  \caption{\textbf{Attention association with explicit occlusion reasoning.}
  Our method explicitly reasons about occlusions by attending to a separate occlusion state.
  The orange circles refer to associated detections, the white circles refer to incoming detections, and the gray circles refer to occlusion states.
  The model classifies a track as occluded if the track embedding most strongly attends to the occlusion state and maintains the state embedding for future association.}
  \label{fig:track}
\end{figure}

\subsection{Attention Association and Explicit Occlusion Reasoning}
\label{sec:ae}


Data association refers to the task of assigning each incoming detection~$d$ in a given frame to an existing track~$T$ or a new track.
To this end, many existing methods estimate a similarity score~$s(d, T)$ between pairs of detections and tracks.
The Hungarian algorithm~\cite{Munkres57} then computes the optimal assignment between detections and tracks using bipartite matching.
In multi-object tracking, however, objects tend to occlude each other, which can lead to missed detections, also known as false negative detections. A robust tracking technique must not erroneously associate another nearby detection with the track that corresponds to the occluded object.

Many methods~\cite{Wojke17,Sadeghian17,Zhu18}, therefore, apply a threshold to the similarity scores to prevent such detections from erroneously being associated with the track.
%
Using the same threshold for all data associations, however, may lead to sub-optimal results despite substantial efforts to fine-tune the threshold, especially on large-scale datasets.

We propose an attention association mechanism to explicitly reason about missed detections.
%
We formulate the data association problem as a dynamic classification problem. Each track effectively chooses one of the detections in the incoming frame. We obtain the classification scores by computing the attention values between a track embedding~$z^T$ and all detection embeddings~$z^d_i$.
In our method, the track embedding~$z^T$ is simply the embedding of the latest associated detection of a trajectory.
%
%
In addition, considering occlusion, a track does not necessarily select one detection out of all available ones, as mentioned earlier in this section.
Therefore, we propose to learn an occlusion state $\text{occ}$ with embedding $z_{\text{occ}} \in \mathcal{R}^{d_k}$, which represents the class that a track can choose in order not to associate with any available detection.
As a result, the probability that a track~$T$ is associated with a detection~$d_i$, or the latent occlusion state~$o$, can be cast as a softmax function, leading to
\begin{equation}
    p(d_i | T) = \dfrac{\exp({{z^d_i}^\top z^T})}{\exp({{z_{\text{occ}}}^\top z^T}) + \sum_j \exp({{z^d_j}^\top z^T)} },
\end{equation}
where the logits are the attention values between the track embeddings and the detection embeddings.
We can then train the model using a cross-entropy association loss for each alive track, giving
\begin{equation}
    \mathcal{L} (T) = -y_{T,o} log(p(o | T)) - \sum_{i} y_{T,i} log(p(d_i | T)) ,
\end{equation}
where $y_{T, i} = 1$ if track~$T$ and detection~$d_i$ belong to the same identity, otherwise $y_{T, i} = 0$,
and $y_{T, o} = 1$ only if $y_{T, i} = 0$ for all $i$.
During inference, each track will associate with the detection $d_i, i= \argmax_{j} p(d_j | t)$.
However, if $p(o | T) > p(d_i | T) $, the track will be marked as ``occluded'' and does not associate with any detection. 

\subsection{Track Management}
We adopt a simple track management technique that is similar to the mechanism used by SORT~\cite{Bewley16} to control the initialization and the termination of tracks.
When an incoming detection is not associated with any track, it is used to initialize an ``unpromoted'' track for which we do not output a tracked target until it is ``promoted``.
An unpromoted track becomes a promoted track when it is associated with any detection in the following frames.
A track will be killed in the system if there is no associated detection in consecutive $T_{lost}$ frames.
We set different thresholds for unpromoted and promoted tracks, denoting as $T_{lost}^{UP}$ and $T_{lost}^{P}$,
where $T_{lost}^{P}$ is usually larger than $T_{lost}^{UP}$ since unpromoted tracks often have a higher probability of containing false positives.
If two tracks compete for one measurement, the optimal assignment, e.g., Hungarian matching, or a greedy algorithm, will assign the track of the higher score to that measurement.
The other track will then typically be classified as occluded because other measurements are usually further from the track and have lower attention values than the occlusion state.
Note that SORT always sets $T_{lost}^{P}=1$ to compensate for a simple motion model.
On the contrary, our proposed method implicitly learns the motion model from data using the proposed attention measurement encoding mechanism. Our technique can further reason about occlusions with the attention association mechanism.
Therefore, both modules enable us to recover from occlusions and maintain tracks longer.

\section{Experimental Evaluation}

We present an extensive experimental evaluation based on several public benchmark datasets. We conduct a detailed ablation study and compare our method with the state of the art in visual multi-object tracking.
%

\subsection{Experimental Setup}

\subsubsection{Implementation Details.}
We implemented the proposed method in Tensorflow. We trained all the models with a single NVIDIA~V100~GPU.
The network is trained with batch size 16 and SGD optimizer with 1e-3 learning rate and 0.9 momentum,
and each training sequence is sampled with 32 consecutive frames with random start timestamp.
The input detection bounding boxes are normalized to $[0, 1]$ with camera width and height.
For simplicity, we set $d_k = 64$ for all fully-connected layers in our network.
We use the track management hyper-parameters $T^{UP}_{lost}=2$ and $T^{P}_{lost}=5$, unless specified otherwise.

\subsubsection{Evaluation Metrics.}
We adopt the CLEAR MOT metrics~\cite{leal2015motchallenge}. These metrics consider the multi-object tracking accuracy~(MOTA), the number of track identity~switches~(IDS), the number of false positives~(FP), and the number of false negatives (FN).
MOTA, however, heavily depends on the performance of the object detector. Therefore, we also include IDF1~score~\cite{ristani2016performance}.
%

\subsubsection{Baseline Methods.}
We compare our approach with the following baseline methods to evaluate different aspects of the techniques:
\begin{itemize}
    \item An \textbf{IOU Tracker} that relies on a constant velocity motion model and uses intersection-over-union~(IOU) as the similarity function between predicted states and detections.
    \item A \textbf{Center Tracker} that relies on a constant velocity motion model and uses negative L2 distance between the box centers of the predicted states and the detections as the similarity function. 
    \item A \textbf{Learned Similarity Tracker} that relies on a learned pairwise similarity function. We encode each detection measurement using 4~fully-connected layers to produce a $64$-d feature for each detection. We compute the cosine similarity between each track embedding~$t$ and detection embedding~$d$ as $t^\top d / (\Vert t \Vert \Vert d \Vert)$. We optimize the similarity using a contrastive loss~\cite{leal2016learning}, where we set the margin to~0.3. We then use Hungarian matching based on the resulting similarity scores.
\end{itemize}


\subsubsection{Public Benchmark Datasets.}
We use the following public benchmark datasets in our experimental evaluation:
\begin{itemize}
    \item The \textbf{Waymo Open Dataset}~\cite{waymo_open_dataset} is a large-scale dataset for autonomous driving. The dataset comprises 798~training sequences and 202~validation sequences. Each sequence spans 20~seconds and is densely labeled at 10~frames per second with camera object tracks.
    We trained and evaluated on the images recorded by the ``front'' camera to track vehicles in our experiments.
    \item The \textbf{KITTI Vision Benchmark Suite}~\cite{REF:Geiger2013IJRR} comprises 21~training sequences and 19~test sequences. We trained on the camera images to track cars in our experiments.
    \item The \textbf{Multiple Object Tracking Benchmark}~\cite{leal2015motchallenge} is a unified framework for evaluating multi-object tracking methods. We adopt the MOT17 benchmark, which comprises 7 sequences featuring crowded scenes with lots of pedestrians.
\end{itemize}

\subsection{Ablation Studies in a Controlled Environment}
\label{sec:controlled_environment}

We first evaluate our approach in a controlled environment that does not depend on the performance of a trained object detector. To this end, we run our method on the ground truth labels of the Waymo Open Dataset as if they were the detections predicted by an object detector.
To simulate occlusions and missed detections, we randomly drop $N_{\text{drop}} \sim U(1,5)$ consecutive bounding boxes for every 10~frames in a given track with probability~$p_{drop}$, processing each ground truth trajectory independently.
We present a performance analysis of the methods in the controlled environment in Table~\ref{table:gt_as_det_drop}.
Both the IOU Tracker and the Center Tracker perform poorly as their constant velocity motion models fail to accurately predict the motion of objects during occlusions.
The Learned Similarity Tracker outperforms both the IOU Tracker and the Center Tracker by a large margin as it is able to recognize previously seen objects based on their appearance.
%
%
%
Our method performs better than all baseline methods.
%
Attention measurement encoding leads to a MOTA improvement of~1.07\,\%~-~1.18\,\% and an IDS reduction of 9.0\,\%~-~9.2\,\%. The explicit occlusion reasoning leads to an even more pronounced IDF1 improvement of~24\,\%.

\begin{table*}[!t]
	\centering
	\caption{
\textbf{Performance in a controlled environment.} We run the methods on the ground truth labels of the Waymo Open Dataset as if they were object detections. Here, we simulate occlusions with $p_{drop}=0.3$.
	}
	\label{table:gt_as_det_drop}
	\small
	\begin{tabular}{lccccccccc}
		\toprule
		Method & AE & Occ  & MOTA $\uparrow$ & IDF1 $\uparrow$ & IDR $\uparrow$ & IDP $\uparrow$ & IDS (k) $\downarrow$ & FP $\downarrow$ & FN $\downarrow$ \\
		\midrule
		IOU &&& 38.0 & 33.7 & 24.0 & 56.6 & 25.2 & 4 & 184.7k\\
		Center &&& 47.9 & 25.2 & 25.1 & 25.4 & 172.1 & 41 & 4.2k \\
		Learned Similarity & & & 83.2 & 30.0 & 29.2 & 30.2 & 50.9 & 15 & 5.5k \\ 
		\midrule
		\multirow{ 4}{*}{Ours} & & & 87.9 & 32.2 & 32.2 & 32.2 & 40.9 & 31 & 60 \\ 
		& \ding{51} && 88.9 & 32.4 & 32.4 & 32.4 & 37.3 &  8 & \bf33 \\ 
		& & \ding{51} & 91.4 & 56.2 & 56.1 & \bf 56.3 & 28.1 & 21 & 1008 \\ 
		& \ding{51} &\ding{51} & \bf 92.7 & \bf 56.3 & \bf 56.2 & \bf 56.3 & \bf 24.0 & \bf 6 & 681 \\ 
		\bottomrule
	\end{tabular}
\end{table*}

\subsection{Evaluation of Explicit Occlusion Reasoning}
\label{sec:occ_exp}
The goal of the explicit occlusion reasoning mechanism introduced in Section~\ref{sec:ae} is to improve the performance when tracking objects that occasionally get occluded.
False positive occlusion predictions, however, may increase the number of missed detections.
Therefore, Table~\ref{table:gt_as_det_drop_occ}~(a) evaluates how the explicit occlusion reasoning performs for different detection drop probabilities~$p_{drop}$.
Interestingly, the two proposed mechanisms affect the tracking quality in different ways depending on the detection noise.
On the one hand, attention measurement encoding leads to the highest gain in MOTA in the absence of simulated occlusions.
On the other hand, attending to the occlusion state leads to the most pronounced improvements when the detections are highly noisy.
In Table~\ref{table:gt_as_det_drop_occ}~(b), we frame the occlusion prediction as a binary classification problem and report metrics, including accuracy, recall, and precision.
The results suggest that our method improves the performance of the occlusion prediction by choosing to operate at a high precision.

%
%

\begin{table}[!t]
    \caption{\textbf{Robustness with respect to missed detections.}
	We evaluate the gain in MOTA achieved by attention measurement encoding~(AE) and explicit occlusion reasoning~(Occ) as a function of the probability of dropped detections. The results in (a) suggest that the explicit occlusion reasoning becomes more beneficial as the rate of occlusions increases. The results in (b) summarize the occlusion classification performance as a function of the drop probability.}
	\label{table:gt_as_det_drop_occ}
	\centering
		\subcaptionbox{\label{}}{
    	\small
    	\begin{tabular}{ccccccc}
    		\toprule
    		Drop prob.& MOTA & Gain. w/ AE & Gain w/ Occ \\
    		\midrule
    		0.0 & 98.6 & 2.1 & -0.5 \\
    		0.1 & 97.0 & 0.8 & 2.0 \\
    		0.2 & 95.4 & 1.6 & 2.5 \\
    		0.3 & 92.6 & 1.0 & 3.6 \\
    		0.4 & 90.8 & 1.1 & 4.8 \\
    		\bottomrule
    		\vspace{1mm}
    	\end{tabular}}

	\subcaptionbox{\label{}}{
		\small
		\begin{tabular}{ccccccc}
    		\toprule
    		Drop prob.& AE & Accuracy $\uparrow$ & Recall $\uparrow$ & Precision $\uparrow$ \\
    		\midrule
    		0.3 &           & 87.4 & 57.6 & 87.1  \\
    		0.3 & \ding{51} & 89.4 & 60.1 & 91.9 \\
    		0.4 &           & 85.6 & 57.9 & 91.7 \\
    		0.4 & \ding{51} & 88.2 & 64.2 & 93.3 \\
    		\bottomrule
    	\end{tabular}
    	}
\end{table}

\subsection{Evaluation of Attention Measurement Encoding}
\label{sec:ae_exp}
\PAR{Hyper-parameters.}
We evaluate the effects of the attention measurement encoding hyper-parameters on the performance in the controlled environment described in~Section~\ref{sec:controlled_environment}.
Specifically, we set~$p_{drop}=0.3$ and consider different values for the size~$L_{enc}$ of the encoding window and the number~$N_{enc}$ of encoding layers.
As shown in Table~\ref{table:gt_as_det_drop_enc_ablation}, as $L_{enc}$ increases, the tracking performance becomes better.
These results suggest that our method leverages all available information from context and history.
We also observe that using an encoding window that spans more than 10~frames only leads to marginal improvements, while introducing more computational complexity.  
Considering the number of stacked encoding layers $N_{enc}$, the performance gain is shown with only 1 layer. When increasing $N_{enc}$, the improvement become more pronounced. However, with $N_{enc}>3$, we observe that it takes the model much longer to converge, while with $N_{enc}=3$, it takes the model only a few epochs. In fact, we found that $N_{enc}>3$ often resulted in unstable states in our experiments.
As a result, we choose $L_{enc}=5$ and $N_{enc}=2$ in all remaining experiments.

\begin{table}[!t]
	\centering
	\small
	\caption{\textbf{Attention measurement encoding hyper-parameters.} 
	We compare the tracking performance in terms of MOTA for different numbers of encoding layers~$N_{\text{enc}}$ as well as different sizes of the encoding window~$L_{\text{enc}}$.}
	\label{table:gt_as_det_drop_enc_ablation}
	\begin{tabular}{cccc}
		\toprule
		 & \multicolumn{3}{c}{$N_{\text{enc}}$} \\ 
		$L_{\text{enc}}$ & \hspace{4mm} 1 \hspace{4mm} & \hspace{4mm} 2 \hspace{4mm} & \hspace{4mm} 3 \hspace{4mm} \\ 
		\midrule
		2 & 89.7 & 89.5 & 89.5 \\
		5 & 90.3 & 91.8 &  92.0 \\
		10 & 91.6 & 92.2 & 92.7 \\
		20 & 92.1 & 92.2 & 92.8 \\
		\bottomrule
	\end{tabular}
	\vspace{-1mm}
\end{table}

\begin{table}[!t]
\centering
	\small
	\caption{\textbf{Improved tracking with future information.} 
	We evaluate the tracking performance when delaying the data association by different numbers $L_{\text{future}}$ of frames.
	The results suggest that the additional implicit state information extracted from future measurements improves the performance of the model. This underlines that it is reasonable to avoid hard data associations when gathering spatiotemporal information.
	}
	\label{table:gt_as_det_drop_future}
	\begin{tabular}{ccccccc}
		\toprule
		$L_{\text{future}}$ & MOTA $\uparrow$  & IDS (k) $\downarrow$  & FP $\downarrow$ & FN $\downarrow$ \\
		\midrule
        0 & 92.5 & 24.5 &  \bf 5 & 676 \\
        2 & 93.3 & 22.0 & 23 & 631 \\
        5 & \bf 93.7 & \bf 20.5 &  7 & \bf 643 \\
		\bottomrule
	\end{tabular}
	\vspace{-1mm}
\end{table}
\PAR{Exploiting future information.}
To demonstrate the benefits of avoiding hard data associations, we conduct experiments in which we delay the data association for $L_{\text{future}}$ frames. In this way, all the detections in time $t$ will be encoded with context information from time $t-L_{\text{enc}}$ to $t+L_{\text{future}}$ with the proposed attention measurement encoding.
We show the results in Table~\ref{table:gt_as_det_drop_future} using ground truth as detection with simulated occlusions where $p_{\text{drop}}=0.3$. Encoding window $L_{\text{enc}}$ and $N_{\text{enc}}$ are set to 5 and 2, respectively.
With only 2 frames of delayed associations, the model outperforms the variants that only have access to past information (see Table~\ref{table:gt_as_det_drop_enc_ablation}).
This implies that the implicit state information of the measurement is improved after integrating more information from the future. Therefore, it is reasonable not to commit to hard associations when gathering spatiotemporal information from past measurements.
The results also demonstrate that our technique performs well when used for offline applications, such as semi-supervised learning.

\subsection{Scalability on Large-Scale Datasets}
\label{sec:scalibility}
\begin{figure}[!t]
  \centering
  \includegraphics[width=0.95\linewidth]{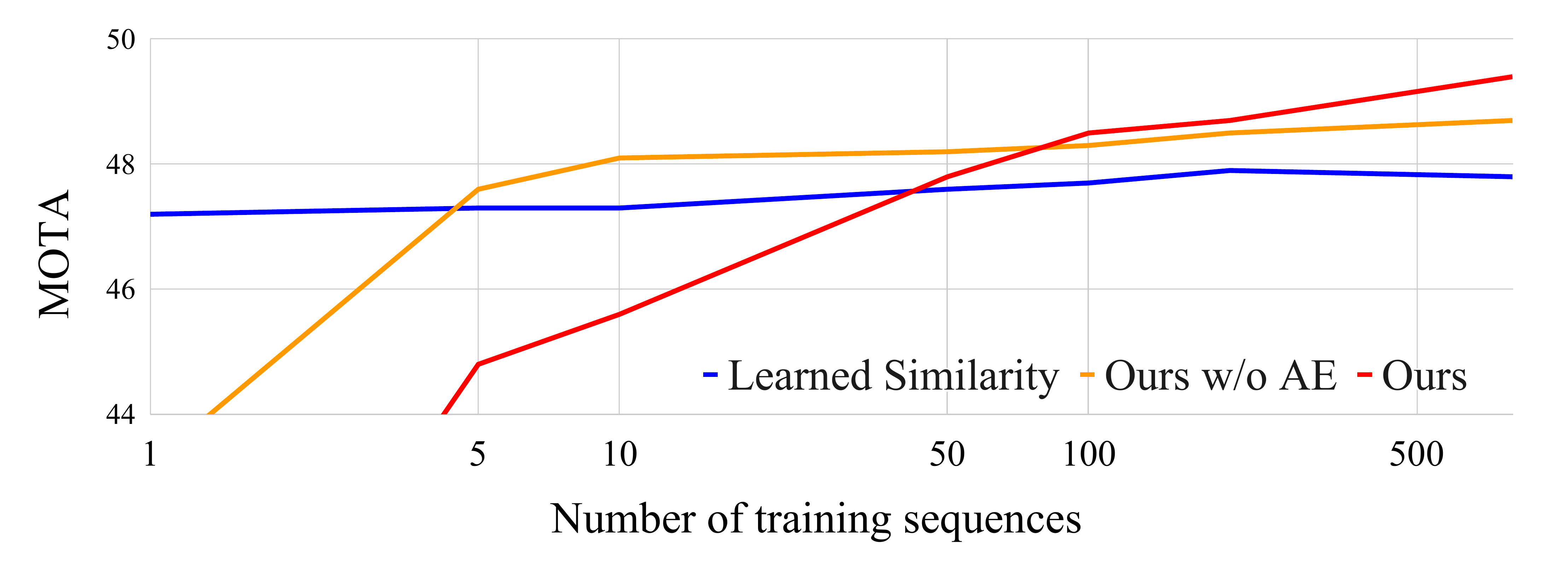}
  \caption{\textbf{Learning curves.} The learning curves suggest that our approach benefits the most from modern large-scale datasets, such as the Waymo Open Dataset.}
  \label{fig:learning_curve}
  \vspace{-3mm}
\end{figure}

The goal of our approach is to learn complicated dependencies between tracks and detections from data.
The learning curves on the Waymo Open Dataset, which are depicted in Figure~\ref{fig:learning_curve}, suggest that our method scales well with the size of the dataset.
As a consequence, our method is able to learn considerably more complex dependencies between tracks and detections from data, which makes it a great technique whenever large amounts of training data are available.

\begin{figure*}[!t]
	\centering
	\begin{tabular}
		{@{\hspace{0mm}}
		c@{\hspace{1mm}} 
		@{\hspace{0mm}}
		c@{\hspace{1mm}}
		@{\hspace{0mm}}
		c@{\hspace{1mm}}
		@{\hspace{0mm}}
		c@{\hspace{1mm}}
		@{\hspace{0mm}}
		c@{\hspace{0mm}}
		}
		\includegraphics[width=0.17\linewidth]{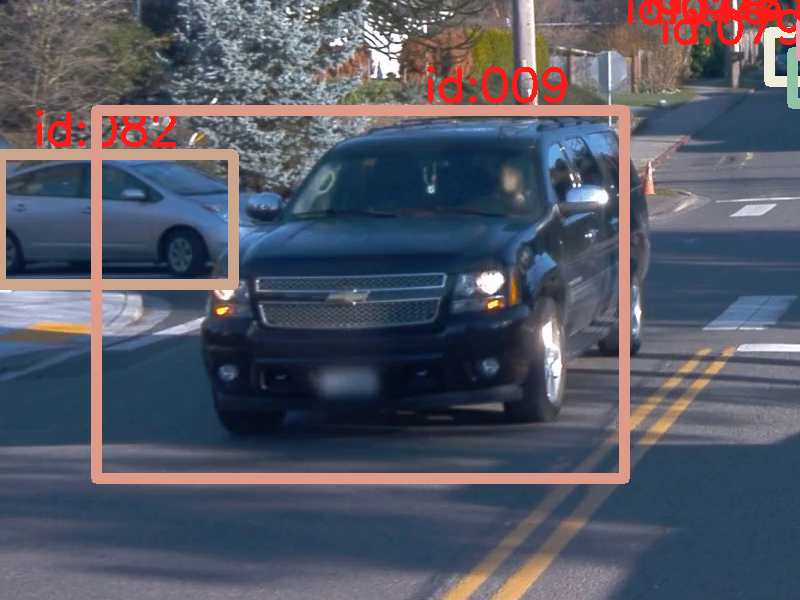} &
		\includegraphics[width=0.17\linewidth]{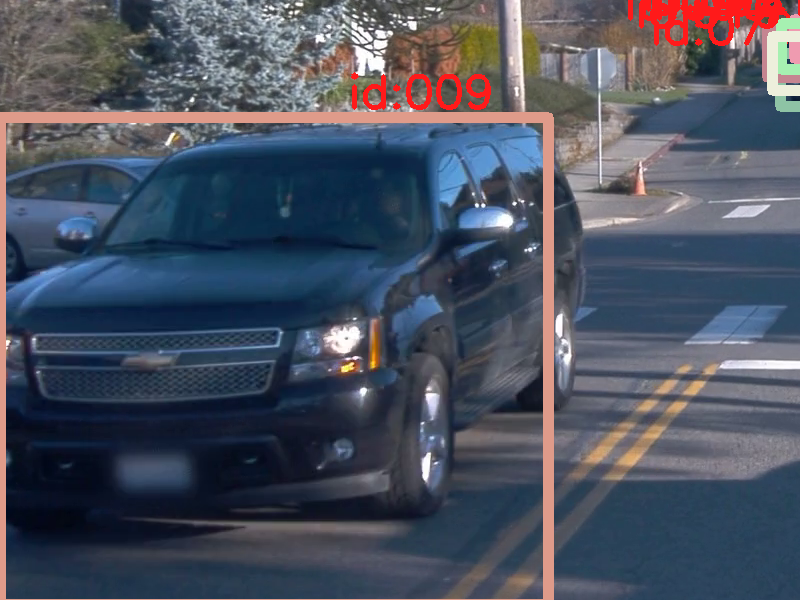} &
		\includegraphics[width=0.17\linewidth]{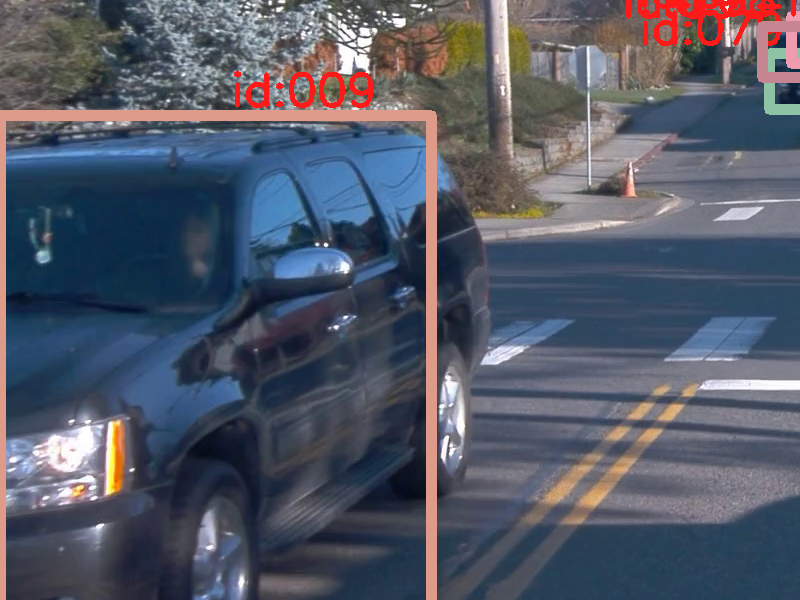} &
		\includegraphics[width=0.17\linewidth]{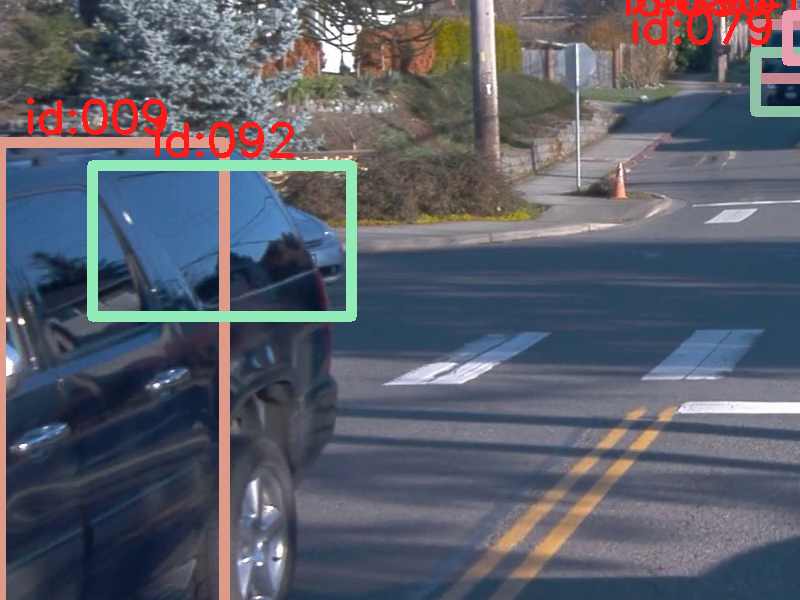} &
		\includegraphics[width=0.17\linewidth]{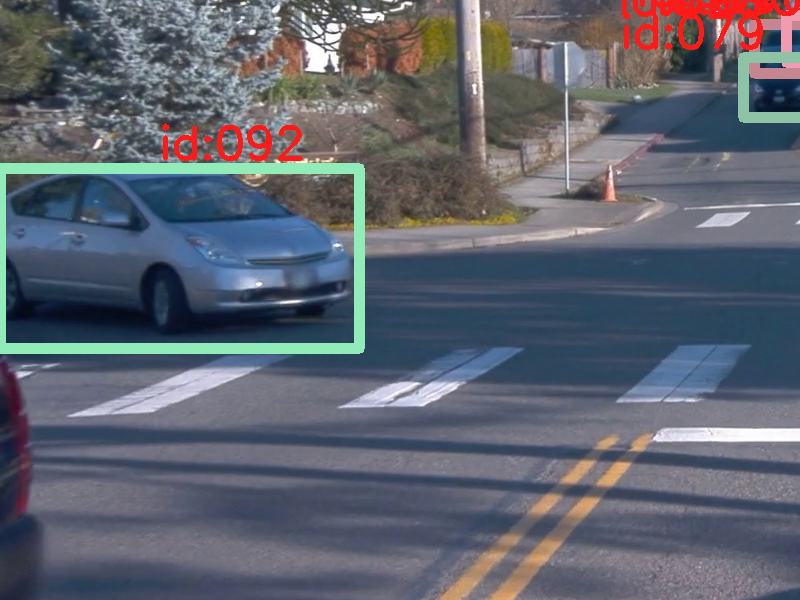} \\
		\includegraphics[width=0.17\linewidth]{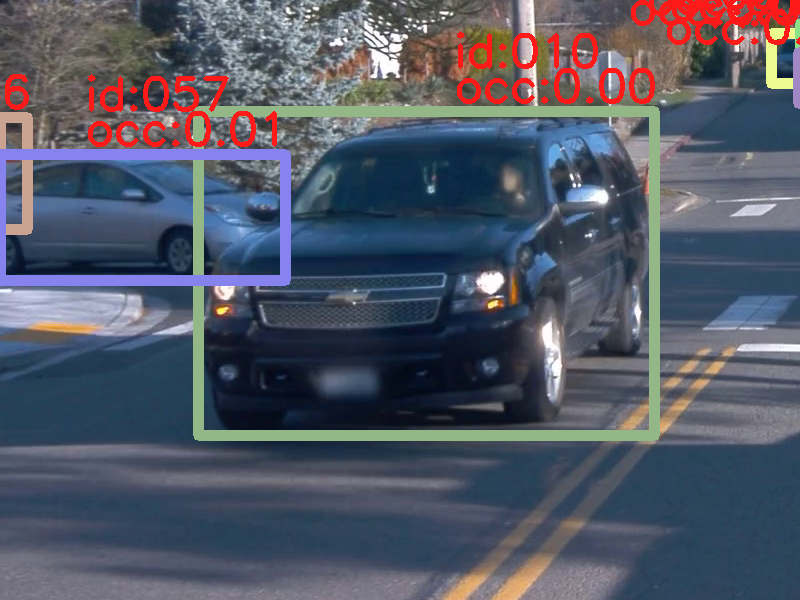} &
		\includegraphics[width=0.17\linewidth]{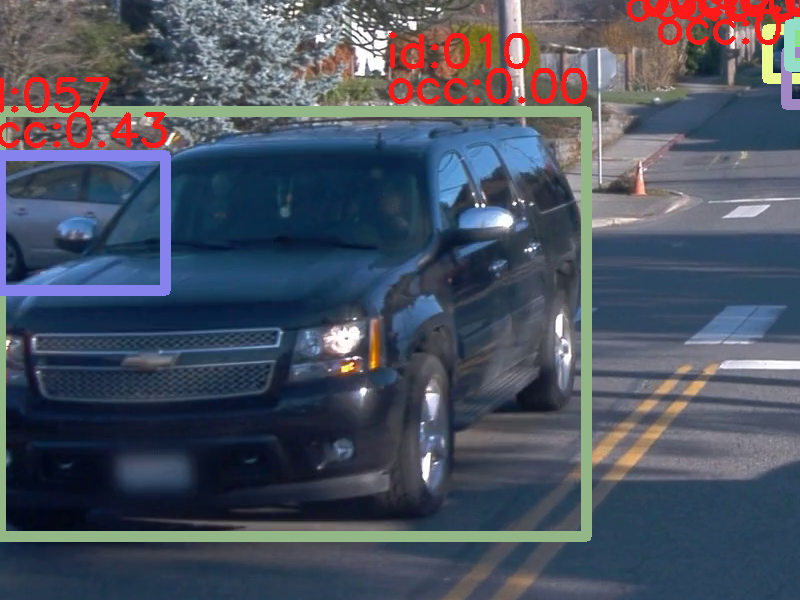} &
		\includegraphics[width=0.17\linewidth]{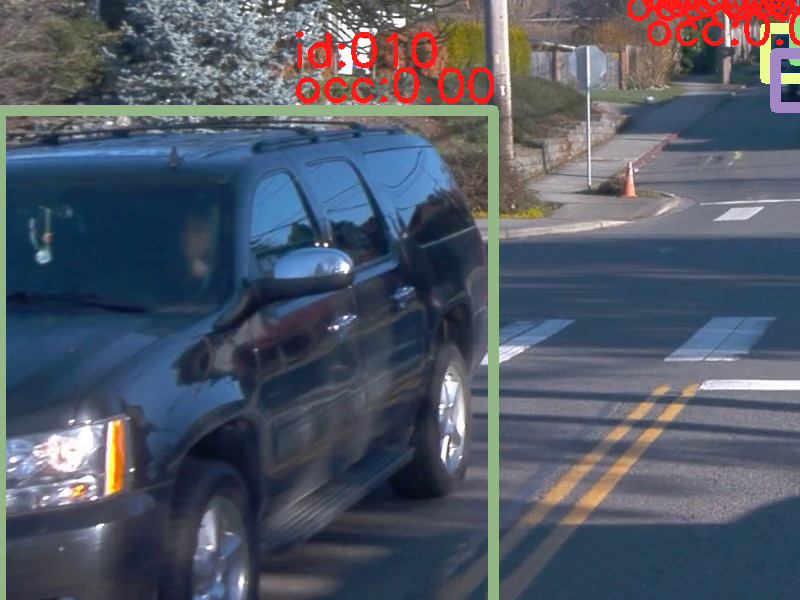} &
		\includegraphics[width=0.17\linewidth]{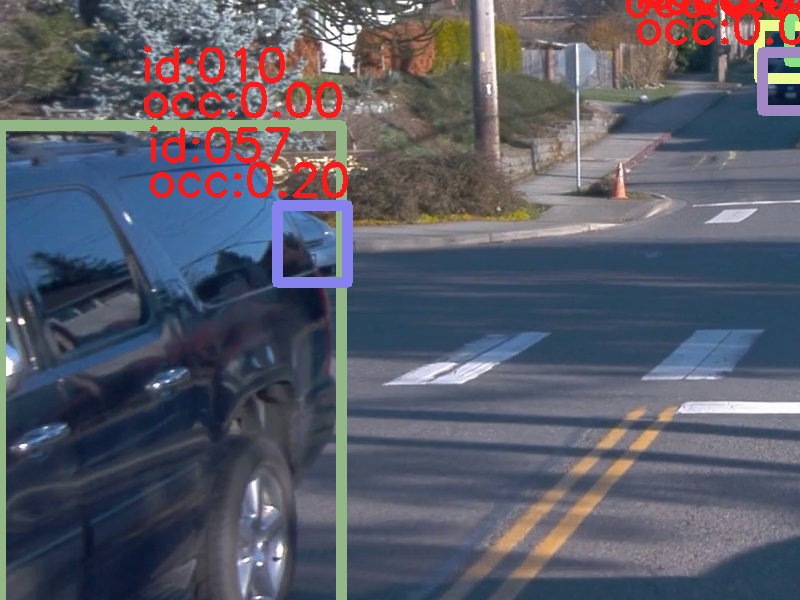} &
		\includegraphics[width=0.17\linewidth]{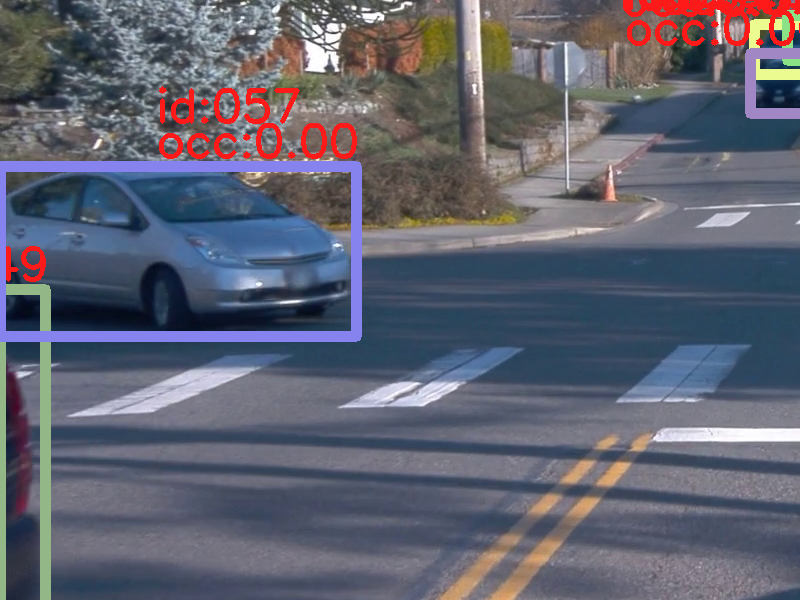} \\
		t = 1 & t = 3 & t=5 & t=7 & t=9
    \end{tabular}
	\caption{\textbf{An example showcasing the explicit occlusion reasoning.}
	We show an occlusion scenario as handled by our model with and without explicit occlusion reasoning. Top:~The baseline model without explicit occlusion reasoning. Bottom: Our method with attention measurement encoding and explicit occlusion reasoning. The car on the left side is occluded between $t=3$ and~$t=7$. At $t=3$, our method attends to the occlusion state with a value of~$0.43$, maintains the track throughout the occlusion, and then recovers the same track at~$t=7$.}
	\label{fig:visual}
	\vspace{-4mm}
\end{figure*}

\subsection{Evaluation on Public Benchmark Datasets}

\subsubsection{Waymo Open Dataset}
We evaluate the proposed method on the Waymo Open Dataset~\cite{waymo_open_dataset} with detections provided by a trained object detector. We compare the method with several baseline methods as well as Tracktor~\cite{Bergmann19}, a state-of-the-art visual tracking method that relies on a two-stage object detector.
To obtain a fair comparison, we train the Faster R-CNN~\cite{ren2015faster} object detector with ResNet-101~\cite{he2016deep} as the backbone network. 
%
%
We refer to the supplementary material for more metrics and training details of the trained detector and the Tracktor method.

We summarize the results in Table~\ref{table:real_detection}~(a).
The IOU Tracker and the Center Tracker both perform poorly as their simple motion models do not accurately capture the complex ego and vehicle motion patterns observed in the scenes.
Tracktor~\cite{Bergmann19} performs~3.2\,\% worse than our Learned Similarity Tracker baseline in terms of MOTA, but~13\,\% better in terms of IDS.
This may be owing to the fact that Tracktor relies on a constant motion model as well as a camera motion compensation to predict the ROI in the next frame. The regression head of the detector is not able to localize objects if the predicted ROI does not have enough overlap with the actual target.
%
%
Tracktor heavily depends on the appearance of objects. In the crowded scenes observed in the Waymo Open Dataset, however, many objects may share similar appearances in nearby locations.

\begin{table*}[!t]
	\caption{\textbf{Performance on public benchmark datasets.}
	}
	\label{table:real_detection}
	\centering
	\small
	\subcaptionbox{Waymo Open Dataset.\label{}}{
	\begin{tabular}{lccccccc}
		\toprule
		Method & MOTA $\uparrow$ & IDF1 $\uparrow$ & IDR $\uparrow$ & IDP $\uparrow$ & IDS (k) $\downarrow$ & FP (k) $\downarrow$ & FN (k) $\downarrow$ \\
		\midrule
		IOU  & 26.5 & 30.1 & {20.3} & 57.8 & 22.6 & \bf 9.6 & 330.2  \\
		Center & 28.7 & 25.7 & 21.1 & 32.6 & 107.5 & 33.7 & 204.1 \\
		Tracktor~\cite{Kim18} & 44.4 & 46.8 & 37.6 & 61.8 & 16.7  & 16.7 & 226.6 \\
		Learned Similarity & 47.8 & 48.6 & 39.9 & 62.1 & 19.1 & 30.1 & 202.0 \\
		\midrule
		Ours w/o AE, Occ & 48.3 & 49.8 & 41.0 & 63.6 & 15.6 & 34.0 & \bf 200.9 \\
		Ours w/o AE  & 49.1 & 55.0 & 45.4 & 65.8 & 13.1 & 32.00 & 201.9 \\
		Ours &  49.4 & \bf 55.8 & \bf 46.0 & \bf 70.9 & 11.4 &  31.9 &  201.9 \\ 
		Ours w/ appearance & \bf 49.5 & 54.1 & 44.3 & 69.4 & \bf 11.3 &  31.9 & 201.8\\ 
		\midrule
		Ours w/ future info &  49.6 & 56.5 &  46.6 &  71.7 & 10.8 & 29.1 & 201.7 \\ 
		\bottomrule
		\vspace{1mm}
	\end{tabular}}
	\subcaptionbox{KITTI-Car Benchmark.\label{}}{
	\label{table:kitti}
	
	\begin{tabular}{lcccccccc}
		\toprule
		Method & MOTA $\uparrow$ & MOTP $\uparrow$ & MT $\uparrow$ & ML $\downarrow$ & FP $\downarrow$ & FN $\downarrow$ & IDS $\downarrow$ \\
		\midrule
		mbodSSP~\cite{lenz2015followme} & 72.7 & 78.8 & 48.8 & 8.7 & 1918 & 7360 & \bf 114 \\
		CIWT~\cite{osep2017combined} & 75.4 & 79.4 & 49.9 & 10.3 & 954 & 7345 & 165 \\
		MDP~\cite{xiang2015learning}& 76.6 & 82.1 & 52.2 & 13.4 & 606 & 7315 & 130\\
		FAMNet~\cite{chu2019famnet} & 77.1 & 79.4 & 49.9 & 10.3 & 954 & 7345 & 165 \\
		BeyondPixel~\cite{sharma2018beyond} & 84.2 & \bf 85.7 & \bf 73.2 & \bf 2.8 & 705 & \bf 4247 & 468 \\
		\midrule
		Ours & 84.2 & 85.3 & 71.1 & 3.3 & 433 & 4531 & 490 \\
		Ours pre-trained on Waymo & \bf 84.3 & 85.3 & 70.3 & 3.5 & \bf 406 & 4575 & 408 \\
		\bottomrule
		\vspace{1mm}
	\end{tabular}}
    \subcaptionbox{MOT17 Benchmark.\label{}}{
	\label{table:mot17}
	\begin{tabular}{lccccccccc}
		\toprule
		Method & \shortstack{Guided \\ Detections} & MOTA $\uparrow$ & IDF1 $\uparrow$ & MT $\uparrow$ & ML $\downarrow$ & \hspace{1mm} FP $\downarrow$ \hspace{1mm} & \hspace{1mm} FN $\downarrow$ \hspace{1mm} & \hspace{1mm} IDS $\downarrow$ \hspace{1mm} \\
		\midrule
		DMAN~\cite{Zhu18} & \ding{51} & 48.2& 55.7 & 19.3 & 38.3 & 26218 & 263608 & 2194 \\
		MOTDT~\cite{chen2018real} & \ding{51} & 50.9 & 52.7 & 17.5 & 35.7 & 24069 & 250768 & 2474 \\
		FAMNet~\cite{chu2019famnet} & \ding{51} & 52.0 & 48.7 & 17.5 & 33.4 & 14138 & 250768 & 2474 \\
		Tracktor++~\cite{Bergmann19}& \ding{51} & 53.5 & 52.3 & 19.5 & 36.6 & 12201 & 248047 & 2072 \\
		\midrule
		Ours & & 43.4 & 30.9 & 15.2 & 35.2 & 21600 & 285129 & 12843 \\
		Ours & \ding{51} & 55.9 & 44.3 & 24.2 & 28.9 & 19683 & 217926 & 11178 \\
		\bottomrule
	\end{tabular}}
	\vspace{-3mm}
\end{table*}

Our method performs favorably compared to the aforementioned methods.
%
Specifically, our method achieves a 5.0\,\% higher MOTA and a 30.8\,\% lower IDS when compared to Tracktor.
%
%
The results suggest that our method is able to account for the noise of the object detector. In addition, the proposed attention measurement encoding effectively exploits spatiotemporal context information to improve data association.
To further evaluate how the attention measurement encoding captures spatiotemporal context information, we evaluate the performance of a variant of our method that has access to the next 2~frames, effectively gathering information from the future. This variant achieves even better performance, suggesting that our method effectively leverages the additional temporal context.
We also evaluate our method when the attention measurement encoding aggregates the appearance features extracted by ROI Align~\cite{he2017mask} along with the detected bounding boxes. The results suggest that the proposed framework generalizes to different feature modalities.
Finally, we present an example of how our method successfully tracks through an occlusion in Figure~\ref{fig:visual}.

\subsubsection{KITTI Tracking Benchmark}
We evaluate the performance of our approach when tracking cars in the camera images. To this end, we use the detections predicted by RRC~\cite{ren2017accurate}. 
The results, summarized in Table~\ref{table:kitti}~(b), suggest that, on this dataset, our method achieves similar performance to the state of the art.
Interestingly, our method achieves even better performance when pre-trained on the Waymo Open Dataset, leading to a $17\,\%$ decrease in IDS. These results confirm the benefits of methods such as ours that leverage modern large-scale datasets. 
Note, however, that we do not use any image features nor 3D priors in this experiment, which is in contrast to other methods, such as BeyondPixel~\cite{sharma2018beyond}.

\subsubsection{MOT17 Benchmark}
As demonstrated in Section~\ref{sec:scalibility}, our method is designed to leverage large-scale datasets comprising hundreds or thousands of scenes, which is in contrast to the MOT17 benchmark.
The results on this dataset, shown in Table~\ref{table:mot17}~(c), are insightful nevertheless.
First, the results suggest that learning an effective association model for the MOT17 benchmark dataset is challenging.
One reason for this may be that the camera view, the frame rate, and the image resolution vary substantially across sequences.
%
It is further worth pointing out that top-performing methods tend to rely on certain guided detections to achieve better performance on this dataset, ranging from highly optimized detectors~\cite{Bergmann19,chen2018real} to single object trackers~\cite{chu2019famnet,Zhu18}.
To obtain the most informative evaluation, we therefore provide results for our method on both the public detections provided by the dataset as well as the private detections used by Tracktor~\cite{Bergmann19}.
%





\section{Conclusion}
\vspace{-0.1cm}
We presented a novel approach to multi-object tracking that leverages the exciting recent developments in attention models in conjunction with the availability of modern large-scale datasets.
We proposed attention measurement encoding to aggregate the rich spatiotemporal context observed in modern datasets into a latent space representation. This allows our model to avoid committing to any hard data associations that may lead to unrecoverable states.
We proposed a mechanism to learn to explicitly reason about occlusions based on the latent space representation. This allows our model to track objects through occlusions while taking into account the context of the scene.
We conducted an extensive experiments on the public benchmark datasets to evaluate the effectiveness of the proposed approach.
The results suggest that our approach performs favorably against or comparable to several baseline methods as well as state-of-the-art methods.
We further demonstrated that our method benefits from large-scale datasets as it is able to learn complex dependencies between tracks and detections from data. 
In future work, we will explore approaches to training our tracking model jointly with the detection model. Furthermore, we will investigate mechanisms to predict refined track estimates based on the spatiotemporal dependencies encoded in the latent space representation.

{\small
\bibliographystyle{ieee_fullname}
\bibliography{references}
}

\appendix
\section{Tracktor Setup on the Waymo Open Dataset}

Since Tracktor~\cite{Bergmann19} only relies on a two-stage detector and a few hyper-parameters, we performed a fair comparison using a vanilla Faster R-CNN~\cite{ren2015faster} with ResNet-101~\cite{he2016deep} as the backbone network.
The model achieves 41.3\% AP (76.7\% $\text{AP}^L$, 42.1\% $\text{AP}^M$, 9.28\% $\text{AP}^S$) on the vehicle class on the Waymo Open Dataset~\cite{waymo_open_dataset}.
To gather the detections for tracking, we set the non-maximum suppression threshold to be 0.6, and we discard all detections with score lower than 0.5.
We optimized the parameters of the Tracktor method on the dataset and set $\sigma_{\text{active}} = 0.4$, $\lambda_{\text{active}} = 0.6$, and $\lambda_{\text{new}} = 0.3$.

\section{Network Architecture Details}

\begin{figure} 
  \centering
  \includegraphics[width=0.6\linewidth]{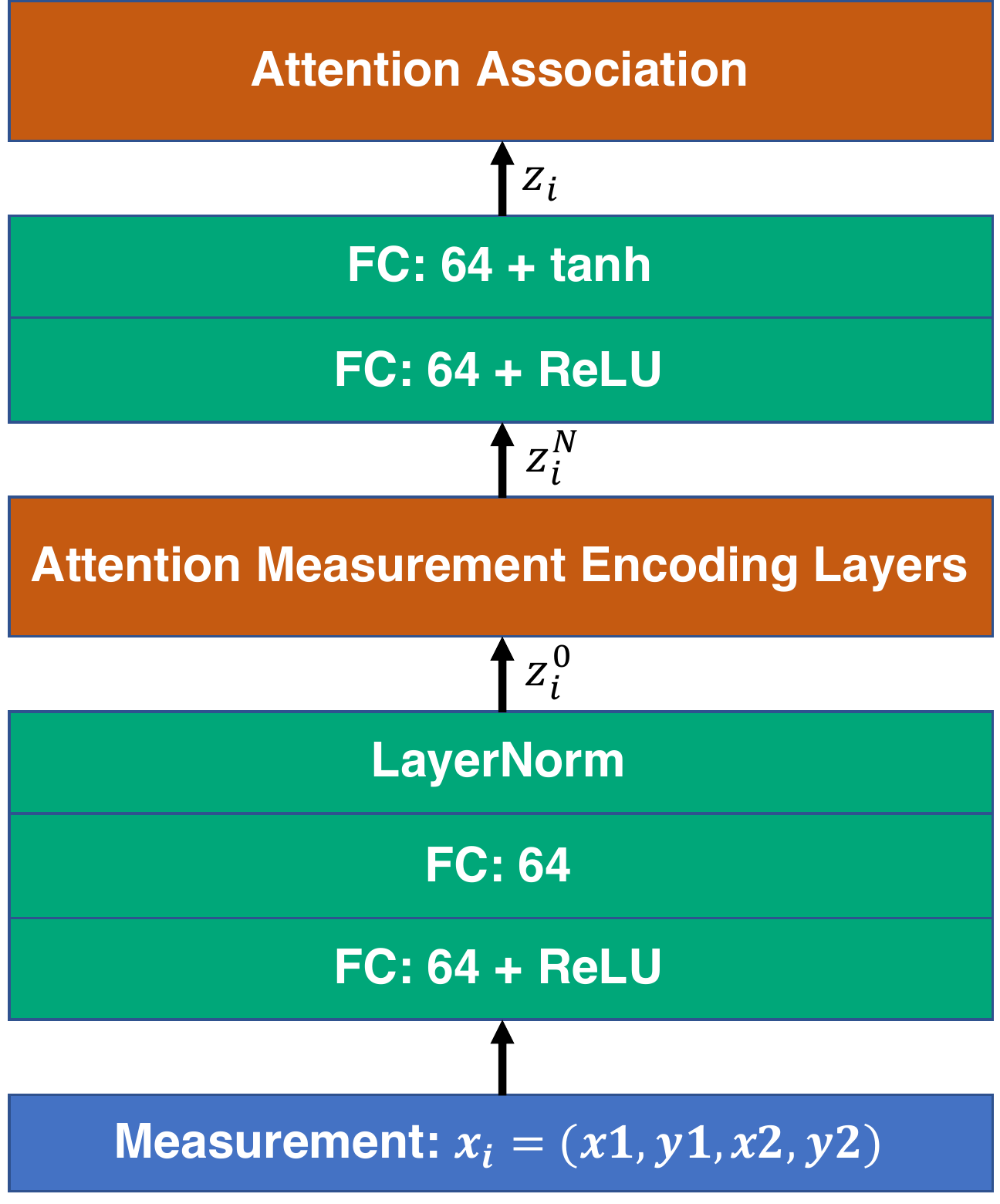}
  \caption{\textbf{Overall Network Architecture.}
  We present the end-to-end network architecture for a detection bounding box. The measurement of a detection $d_i$ is denoted as $x_i$. The embeddings before and after $N$ attention measurement encoding layers are denoted as $z_i^0$ and $z_i^N$, respectively. Before the attention association, we apply two more layers to obtain the final embedding $z_i$.}
  \label{fig:arch}
\end{figure}

We show the detailed network architecture of the proposed method in Figure~\ref{fig:arch}. For all the experiments presented in the paper except \textit{Ours w/ appearance} in Table 5 of the paper, we only use the bounding box coordinates $(x_1,y_1,x_2,y_2)$ as the detection measurement values. Before the attention measurement encoding layers, the measurements are passed through two fully-connected layers and LayerNorm~\cite{ba2016layer}. After encoding with the spatiotemporal context information, the embeddings are further passed through two fully-conntected layers. However, the activation of the final layer is \textit{tanh} since we found it leads to a more stable training process.

\section{Experiments with Simulation Dataset}

\begin{figure}
  \centering
  \includegraphics[width=0.9\linewidth]{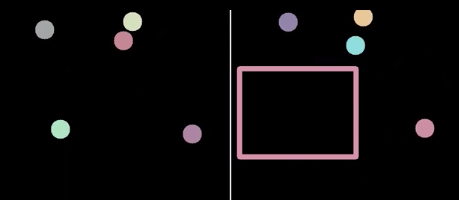}
  \caption{\textbf{Simulation Environment Illustration.} To better understand the challenges introduced by different noise and group dynamic models, we create a simulation environment and separately impose several types of challenges, including measurement noise, mutual/environmental occlusions, and group context. Left: The true state of the objects. Right: Simulated detections affected by occlusions and position noise.
  }
  \label{fig:synthetic}
\end{figure}

We create a simulated environment to evaluate the proposed multi-object tracking method with different challenges that could be faced in MOT.
In the simulation, we put $N_p$~particles of the same size in the box with random initial positions~$p = (p_x, p_y) \sim U(0,1)$ and velocities~$v_i = (v_x, v_y) \sim N(0, 0.1)$.
In each time step, we compute a new position of a particle as~$p^{t+1} = p^{t} + v{t}$.
When the center of a particle exceeds the boundary, it will bounce back with reversed velocity $(-v_x, v_y)$ or $(v_x, -v_y)$. 
To obtain the detection results, we add another random perturbation to the groundtruth position $d = p + n_d, n_d \sim N(0, 0.05)$, serving as the
input to tracker.
Since we focus on learning the motion and context modeling in this work, the appearance feature is not considered in this setting.
For each setting, we generate 1000 train sequences and 20 test sequences. Each sequence contains 600 frames.
We illustrate a sample image for both generated GT and detections in Figure~\ref{fig:synthetic}.

\PAR{Basic Environment with Detection Noise.}
We first evaluate the proposed method in the simplest setup, where there is only displacement noise $n_d$ in the simulation.
We further apply a random force $f = (f_x, f_y) \sim N(0, 0.01) $ to the particles to simulate independent environmental forces.
We report the quantitative evaluation in Table~\ref{table:syn_1} with $N_p=5$.
We first show the performance of a standard Kalman filter-based tracker denoted as \textit{IOU tracker}.
Owing to the simplicity of the dataset, the IOU tracker can already achieve $94.08\%$ MOTA.
The \textit{Learned} baseline performs slightly better than the IOU tracker with $0.54\%$ in MOTA.
However, the number of id switches is higher than that of the IOU tracker while leading to fewer false negatives because a learned tracker could still associate detections that have no overlap with previous bounding boxes.
We show the performance of the proposed method in Table~\ref{table:syn_1} with the ablation study of two proposed components:
\textbf{AE} represents attention measurement encoding, and \textbf{Occ} represents the existence of the occlusion state for attention association.
%
%
All the variants of our method achieve higher MOTA than the baselines, while the attention measurement encoding improves the tracking performance with a $5\%$ reduction in the id switches
with or without occlusion reasoning.
We note that the occlusion state does not improve the tracking performance much since there is no actual occlusion in this simplest setting.

\begin{table}[!t]
	
	\small
	\centering
	\begin{tabular}{ccccccc}
		\toprule
		Method & AE & Occ & MOTA & IDS & FP & FN \\
		\midrule
		IOU Tracker & & & 94.08 & 1642 & 740 & 1168  \\
		Learned & & & 94.62 & 1731 & 745 & 745 \\ 
		\midrule
		\multirow{ 4}{*}{Ours} & & & 94.84 & 1669 & 732 & 732 \ \\ 
		& \ding{51} & & 94.96 & 1589 & 740 & 740 \\ %
		& & \ding{51} & 94.65 & 1671 & 736 & 736 \\ %
		& \ding{51} &\ding{51} & \bf 95.07 & \bf 1573 & \bf 731 & \bf 731 \\ %
		\bottomrule
	\end{tabular}
	\vspace{0.1cm}
	\caption{\textbf{Track with random force and measurement noises.} We show the comparisons between the baseline methods and the proposed method on the synthetic dataset with radom forces and measurement noises. The results show that the proposed method can better understand the motion model of the particles with attention measuremetn encoding.}
	\label{table:syn_1}
\end{table}

\begin{table}[]
	\small
	\centering
	\begin{tabular}{ccccccc}
		\toprule
		Method & AE & Occ & MOTA & IDS & FP & FN \\
		\midrule
		IOU tracker & & & 75.59 & 2321 & 417 & 11907 \\
		Learned & & & 91.59 & 1962 & 511 & 2563 \\ %
		\midrule
		\multirow{ 4}{*}{Ours} & & & 91.82 & 1818 & 518 & 2558 \ \\ %
		& \ding{51} && 91.90 & 1793 & \bf 507 & 2550 \\ %
		& & \ding{51} & 91.86 & 1800 & 515 & 2556 \\ %
		& \ding{51} &\ding{51} & \bf 91.95 & \bf1752 & 511 & \bf 2550 \\ %
		\bottomrule
	\end{tabular} 
	\vspace{0.1cm}
	\caption{\textbf{Track with Occlusions.}
	We evaluate the proposed method on the synthetic dataset with simulated occlusions: mutual and environmental occlusion. The results show that the explicit occlusion reasoning (Occ) is able to reason about occlusion explicitly and recover the track when detection shows up again.}
	\label{table:syn_2}
\end{table}

\begin{table*}[!t]
	\small
	\centering
	\begin{tabular}{ccc|cccc|cccc}
		\toprule
		 & & & \multicolumn{4}{c|}{$N_p=5$} & \multicolumn{4}{c}{$N_p=10$} \\
		Method & AE & Occ & MOTA & IDS & FP & FN & MOTA & IDS & FP & FN \\
		\midrule
		IOU tracker & & & 94.7 & 2809 & 113 & 234 & 86.6 & 14.6k & 568 & 871\\
		Learned & & & 95.2 & 2634 & 123 & 123 & 88.0 & 12.6k & 257 & 257\\
		\midrule
		\multirow{ 4}{*}{Ours} & & & 95.3 & 2606 & 116 & 116  & 87.5 & 13.8k & 605 & 605\\ %
		& \ding{51} && \bf 97.9 & \bf 1130 & \bf 66 & \bf 66 & \bf 93.4 & \bf7.1k & \bf 446 & \bf 446\\ %
		& & \ding{51} & 95.3 & 2605 & 114 & 114 & 87.3 & 14.1k & 597 & 597\\ %
		& \ding{51} &\ding{51} & 97.7 & 1204 & 74 & 74 & 93.3 & 7.1k & 449 & 449\\ %
		\bottomrule
	\end{tabular} 
	\vspace{0.1cm}
	\caption{\textbf{Track with Social Forces.}
	We evaluate the proposed method on the synthetic dataset with social forces. 
	The results demonstrate that the proposed attention measurement encoding (AE) could leverage the context information from other particles to better track the targets.}
	\label{table:syn_3}
\end{table*}

\PAR{Track with Occlusions.}
To simulate challenging cases in multi-object tracking, we inject occlusion noise into the simulation.
Specifically, we simulation two types of occlusions: mutual occlusions and environmental occlusions.
To simulate mutual occlusions, we assign a random depth value to each particle.
When the IOU between two particles is higher than $0.3$, we remove the detection of the particle with larger depth value.
For environmental occlusion, we simulate an occlusion block with random position and size in each sequence.
Every particle that has overlap will be marked missed in the detections.

We apply the baseline methods and our approach to this synthetic occlusion dataset and show the comparisons in Table~\ref{table:syn_2}.
With synthesized occlusion, the performance of IOU tracker drops drastically compared to Table~\ref{table:syn_1} because it is not able track the particles correctly after occlusion, resulting many false negatives and id switches.
Learned trackers, however, suffer less from the additional occlusion noise.
With the proposed method, the attention measurement encoding, comparing to occlusion-free environment, still bring improvement with 2.5\% drop of id switches.
On the other hand, the explicit reasoning of occlusion state improve the id switches by 2.7\% since there is systematic occlusion noise introduced.

\PAR{Track with Social Forces.}
Finally, to demonstrate the effectiveness of the proposed attention measurement encoding to learn from spatiotemporal context information, we apply social forces~\cite{helbing1995social} to the particles by considering the \textit{repulsive effects} between the particles, where they try to not collide with each other. Note that there is no occlusion simulation.
Specifically, let $p_i$ be the position of a particle~$i$, the repulsive social forces for the specific particle $i$ can be formulated as 
\begin{equation}
    f_i = \sum_j F_0 e^{-\frac{||p_i - p_j||}{R}} \cdot \nabla (p_i - p_j),
\end{equation}
where $j$ denote the index of all the other particles in the scene, $F_0$ denotes the base force magnitude,$R$ denotes the tolerance radius, and $\nabla (p_i - p_j)$ denotes the unit vector from $j$ to $i$. By simulating the simple social force model, the particles movements are now depending on each other, and therefore the state of the other particles would be an important information for reliably tracking a particle.

We show the comparisons in Table~\ref{table:syn_3} with both $N_p=5$ and $N_P=10$ settings to illustrate how the trackers performs in different densities.
Without any occlusions, the IOU tracker performs reasonably well. However, id switches occur when particles interact with each other and change their motion drastically because of that.
Similar results are shown in both the \textit{Learned} baseline and our methods baseline since it is difficult to track the objects without the context information.
However, with the proposed attention measurement encoding, the MOTA is improved by $2.4\%$ with $N_p=5$ and $5.9\%$ with $N_p=10$, and the number of id switches are greatly reduced.
This is because the tracker is able to reason about the interaction between particles and associate the particles with better state estimation.

\end{document}